\documentclass[10pt,twocolumn,letterpaper]{article}

\usepackage{iccv}
\usepackage{times}
\usepackage{epsfig}
\usepackage{graphicx}
\usepackage{amsmath}
\usepackage{amssymb}

\usepackage[pagebackref=true,breaklinks=true,letterpaper=true,colorlinks,bookmarks=false]{hyperref}

\iccvfinalcopy % *** Uncomment this line for the final submission

\def\iccvPaperID{10392} % *** Enter the ICCV Paper ID here

\ificcvfinal\fi

\usepackage[utf8]{inputenc} % allow utf-8 input
\usepackage[T1]{fontenc}    % use 8-bit T1 fonts
\usepackage{hyperref}       % hyperlinks
\usepackage{url}            % simple URL typesetting
\usepackage{booktabs}       % professional-quality tables
\usepackage{amsfonts}       % blackboard math symbols
\usepackage{nicefrac}       % compact symbols for 1/2, etc.
\usepackage{microtype}      % microtypography

\usepackage{times}
\usepackage{epsfig}
\usepackage{graphicx}
\usepackage{amsmath}
\usepackage{amssymb}
\usepackage{bm}
\usepackage{caption}

\newcommand{\xx}{\mathbf{x}}

\newcommand{\rr}{\mathbf{r}}
\newcommand{\ttt}{\mathbf{t}}

\newcommand{\cL}{\mathcal{L}}

\newcommand{\BB}{\mathbf{B}}

\newcommand{\WW}{\mathbf{W}}
\newcommand{\VV}{\mathbf{V}}
\newcommand{\MM}{\mathbf{M}}
\newcommand{\JJ}{\mathbf{J}}
\newcommand{\CC}{\mathbf{C}}
\newcommand{\EE}{\mathbf{E}}

\usepackage{xspace}
\makeatletter
\DeclareRobustCommand\onedot{\futurelet\@let@token\@onedot}
\def\@onedot{\ifx\@let@token.\else.\null\fi\xspace}

\def\ie{\emph{i.e}\onedot}

\makeatother

\begin{document}

%%%%%%%%% TITLE
\title{THUNDR: Transformer-based 3D HUmaN  Reconstruction with Markers}

\author{%
  Mihai Zanfir \\
  \texttt{mihaiz@google.com} \\
  \and
 Andrei Zanfir \\
  \texttt{andreiz@google.com} \\
  % examples of more authors
  \and
  Eduard Gabriel Bazavan \\
  % Affiliation \\
  % Address \\
  \texttt{egbazavan@google.com} \\
  \and
  William T. Freeman \\
  \texttt{wfreeman@google.com} \\
  \and
  Rahul Sukthankar \\
  \texttt{sukthankar@google.com} \\
  \and
  Cristian Sminchisescu \\
  \texttt{sminchisescu@google.com} \\
  \\
  {\bf \large Google Research}\\
}

\twocolumn[{%
\renewcommand\twocolumn[1][]{#1}%
\maketitle
\begin{center}
    \centering
    \includegraphics[width=.9\textwidth]{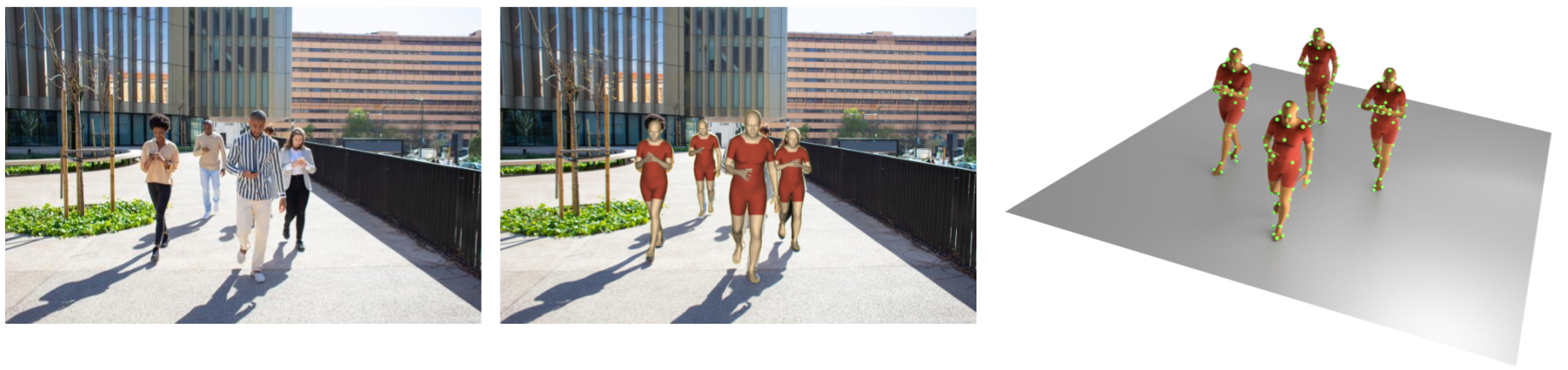}
    \vspace{-4mm}
    \captionof{figure} {Automatic 3d pose and shape reconstruction results with THUNDR. \textit{(Left)} Input image. \textit{(Middle)} Reconstructed 3d meshes projected on the camera plane and overlayed on the image. \textit{(Right)} Different viewpoint showing our intermediate predicted marker representation (in green) and the reconstructed surface geometry. THUNDR provides automatic 3D scene placement of the reconstructed humans under a perspective camera model.}
\end{center}%
}]
\ificcvfinal\thispagestyle{empty}\fi

%%%%%%%%% ABSTRACT
\begin{abstract}
We present THUNDR, a transformer-based deep neural network methodology to reconstruct the 3d pose and shape of people, given monocular RGB images. Key to our methodology is an intermediate 3d marker representation, where we aim to combine the predictive power of model-free-output architectures and the regularizing, anthropometrically-preserving properties of a statistical human surface model like GHUM---a recently introduced, expressive full body statistical 3d human model, trained end-to-end. Our novel transformer-based prediction pipeline can focus on image regions relevant to the task, supports self-supervised regimes, and ensures that solutions are consistent with human anthropometry. 
%As a by-product, we show how marker data can be easily converted into a complete parametric human model (in our case GHUM), by employing a novel deep learning architecture. 
We show state-of-the-art results on Human3.6M and 3DPW, for both the fully-supervised and the self-supervised models, for the task of inferring 3d human shape, joint positions, and global translation. Moreover, we observe very solid 3d reconstruction performance for difficult human poses collected in the wild.
\end{abstract}

%%%%%%%%% BODY TEXT
\section{Introduction}

The significant recent progress in 3d human sensing is supported by the development of statistical human surface models 
%like SMPL or GHUM, 
and the emergence of different forms of supervised and self-supervised visual inference methods.
%like SPIN, Holopose, or HUND, among others. 
The use of statistical human pose and shape models offers advantages in the use of an anatomical and semantically meaningful human body representation, during both learning and inference. Human anthropometry could be used to regularize a learning and inference process, which, in the absence of such constraints, and given the ambiguity of 3d lifting from monocular images, could easily run haywire. This is especially true for unfamiliar and complex poses not previously seen in a `training set'---as they never all are. Semantic models offer not only correspondences with image detector responses (specific body keypoints or semantic segmentation maps) which can give essential alignment signals for 3d self-supervision, but can also help rule out 3d solutions that may otherwise entirely break the symmetry of the body, the relative proportions of limbs, the consistency of the surface in terms of non self-intersection, or the anatomical joint angle limits. 

The choice of evaluation metrics has an important role, too. For now, by far the most used representation---perceived as `model-independent'---are the `body joints', a popular concept, neither by virtue of its anatomical clarity (as that point idealization could be bio-mechanically argued against), nor---for computer vision, and more practically---given its lack of ground-truth observability. In practice, human `body joints' are obtained either by fitting proprietary articulated 3d body models to marker data (internal models of the Mocap system, where the assumptions and error models are not always available) or by human annotators eye-balling joint positions in images, followed by multi-view triangulation to obtain pseudo-ground truth. While the latter have proven extremely useful in bootstraping initial 3d predictors, the joint-click positioning cannot be considered an accurate anatomical reality, in any single image, and even less so, consistently, over a large corpora, especially as for many non-frontal-parallel  poses `joint locations' are difficult to correctly identify, visually. 
%Overall, it seems unwise to rely on metrics so heavily dependent on never-observed structures, and with uncertain and possibly inconsistent identification protocols. This is especially true when evaluating papers and methods in a relatively narrow band of performance, deciding acceptance or the state of the art. 
While some form of 3d body joint prediction error seems unavoidable under the current ground-truth and state of the art metrics, a safeguard could be to operate primarily with visually grounded structures and obtain joint estimates using statistical body models, based on their surface estimates, as just a final step. 
%The question is how to combine the advantages of both worlds.
%to at least model t and primarily operate with visually grounded structures.

In this paper, we rely on the visual reality of 3d body surface markers (in some conceptualization, a `model-free' representation) and that of a 3d statistical body (a `model-based' concept) as pillars in designing a hybrid 3d visual learning and reconstruction pipeline. Markers have the additional advantages of being useful for registration between different parametric models, can be conveniently relied-upon for fitting, and can be used as a reduced representation of body shape and pose, as we will here show. Our model combines multiple novel transformer refinement stages for efficiency and localization of key predictive features, and relies on combining `model-free' and `model-based' losses for both accuracy and for results consistent with human anthropometry. Quantitative results in major benchmarks indicate state of the art performance. Extensive qualitative testing in the wild supports the overall feasibility, and the quality of 3d reconstructions produced by THUNDR, under both supervised and self-supervised regimes.

\noindent{\bf Related Work:} There is considerable prior work in 3d human sensing which we only briefly mention here without aiming at a full literature review \cite{sminchisescu_ijrr03,bogo2016,SMPL2015,ghum2020,dmhs_cvpr17,zanfir2018monocular,Rhodin_2018_ECCV,Kanazawa2018,kolotouros2019learning,ExPose:2020,choi2020pose2mesh,zanfir2020neural}. Methods sometimes referred as `model-based' \cite{zanfir2018monocular,kolotouros2019learning,guler2019holopose,moon2020MeshNet,jiang2020coherent,biggs20203d,xu2019denserac,zhang2020object,arnab2019exploiting,zeng20203d,georgakis2020hierarchical} rely on statistical human body models like SMPL or GHUM, whereas others sometimes referred to as `model-free' \cite{varol18_bodynet,sun2018integral,iqbal2020weakly,zeng20203d} rely on predicting a set of markers or mesh positions, without forms of statistical surface or kinematic regularization based on human anthropometry. While the second class of techniques tend to perform better in benchmarks (which are mostly emphasizing the prediction of 3d joint locations and occasionally joint angles), the former tend to be more semantically and anatomically intuitive, easier to deploy in the context of self-supervised learning, and more robust in environments, or for poses, not encountered during training. In this work we aim to leverage the advantages of both methods: predicting visually observable sets of markers, and yet regularize estimates using statistical kinematic pose and shape models. Moreover, additional innovations in the use of multiple layers of refining visual transformers, produce significant computational efficiency and accuracy gains in benchmarks, for self-supervised learning, and in the wild.

\section{Methodology}

In this section we review our methodology including the 3d statistical body models, the marker based-modeling, as well as the proposed THUNDR learning and inference architecture.

\subsection{Statistical 3D Human Body Models}
\label{subsection:GHUM_CAMERA}
We use a recently introduced statistical 3d human body model called GHUM \cite{ghum2020}, to represent the pose and the shape of the human body. The model has been trained end-to-end, in a deep learning framework, using a large corpus of human shapes and motions. 
%and different from subjects S9 and S11 used for testing).  
The model has generative body shape and facial expressions $\beta = \left( \beta_b, \beta_f \right) $ represented using deep variational auto-encoders and generative pose $\theta = \left( \theta_{b}, \theta_{lh}, \theta_{rh} \right)$ for the body, left and right hands respectively represented as normalizing flows \cite{zanfir2020weakly}. The pelvis translation and rotation are controlled separately, and represented by a 6d rotation representation \cite{zhou2018continuity} $\rr\in\mathbb{R}^{6\times1}$ and a translation vector $\ttt\in\mathbb{R}^{3\times1}$ w.r.t the origin $(0, 0, 0)$. The mesh consists of of $N_v = 10,168$ vertices and $N_t = 20,332$ triangles. To pose the mesh, we apply the GHUM network $\mathbf{V}(\theta_{b}, \beta_{b}, \rr, \ttt) \in \mathbb{R}^{N_v \times 3}$ to obtain the posed vertices. We omit the facial expressions and left and right hand poses, as we here focus on main body pose and shape. We also drop the $b$ subscript for convenience.

\vspace{-5mm}
\paragraph{Camera Model} We adopt a pinhole camera model, with intrinsics $\CC = [f_x, f_y, c_x, c_y]^\top$ and associated perspective projection operation $\xx_{2d} = \Pi(\xx_{3d}, \CC)$, where $\xx_{3d} \in \mathbb{R}^{3\times1}$. Because we work with cropped images, we also adapt our intrinsics, such that projecting the same 3d points -- either in the cropped image or the original, full image -- would give the same alignment.
The transformation of image intrinsics $\CC$ into corresponding crop intrinsics $\CC_c$ is given by
\begin{align}
    [\CC_c^\top 1]^\top = \mathbf{K} [\CC^\top 1]^\top,
\end{align}
where $\mathbf{K} \in \mathbb{R}^{5\times 5}$ is the scale and translation matrix, adapting the image intrinsics $\CC$. 

By using a perspective camera model, we ensure that reconstructions are obtained in camera space. Hence, we have meaningful translation and relative positioning of one subject in the scene (or relative positioning of multiple subjects) when reconstructing from monocular images. A perspective model is a much more accurate and general representation of the imaging transformation compared to an orthographic one. It is in our view desirable in all cases, in order to go beyond showing just model projections or reconstructions in a human-centred coordinate system.
%This is also useful in
%where exact localization in a 3d scene is needed---arguably in order to properly place the subject and track their global motion in depth, and 
%capturing perspective distortion.

\subsection {Marker-based Modelling}
\label{subsection:MARKER_POSER}
Current model-based architectures directly predict specific shape or pose parameters from a raw image. Inspired by model-free methods where weaker constraints are applied on outputs, we adopt an intermediate representation given by 3d surface markers. These can capture human shape and pose and we can predict them directly in a 3d camera space. 

However, training purely model-free methods based on surface markers (as opposed to joint locations), faces additional challenges for both supervised and unsupervised learning. First as such markers are different from joint positions, very few datasets have labels for them. Markers may be available for motion capture datasets in both 2d and 3d, but training a reliable detector is not necessarily easy especially if one seeks generalization outside the lab, where most marker-based systems operate. For self-supervised learning, where additional forms of semantic (body part segmenation) analysis are often necessary, the lack of a statistical body model would render such potentially useful signals unavailable. Finally---and especially when learning with small supervised training sets or for exploratory self-supervised learning---, the lack of regularization given by a body model could lead to 3d predictions with inconsistent anthropometry, further derailing a convergent learning process.

%2d labels only, or even 3d -- where no 3d marker information is available --, markers cannot be used to minimize any meaningful loss because: i) a reprojection loss required 2d marker positions, and there no current architectures or datasets that predicts or labels them, ii) for semantic-alignment metric you need a full body mesh, iii) even if such data would exist, there should be antropometrical constraints placed -- such as bone lengths, angle limits etc. -- that are not straightforward to set with markers. For these reasons, we also want to have a function that, given a set of 3d markers, outputs the mesh that is described by those markers. This only solves just half of the problem -- we can get 2d or 3d keypoints and semantic body-part labels from that mesh -- but we still need to impose antropometrical constraints. Therefore, we would also want a function that would output the generative codes for body shape and pose that would best fit a GHUM mesh that describe the markers. We can then assume a separable prior $p(\thetab,  \betab)=p(\thetab) + p(\betab)$ where Gaussian components with $\mathbf{0}$ mean and unit $\mathbf{I}$ covariance, that can be used to place antropometric constraints on the marker reconstructions.

\begin{figure*}[!tbp]
\begin{center}
\includegraphics[width=.9\linewidth]{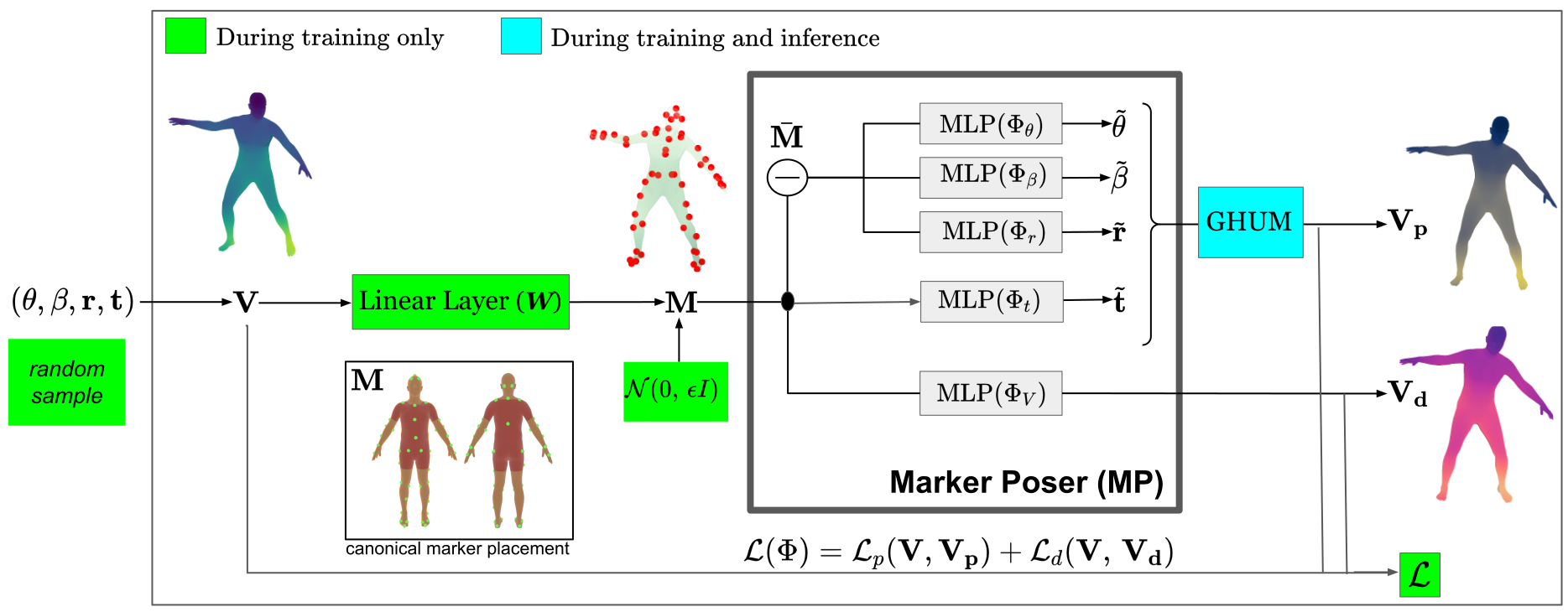}
\end{center}
\vspace{-5mm}
\caption{\small Our marker poser is based on a constrained marker-prediction pipeline which auto-encodes an initially generated, body mesh that is consistent with the human anthropometry ${\bf V}$ into a set of markers $\MM$ via a linear layer characterized by a matrix $\WW$. The markers are then used to predict both the GHUM parameters, resulting in a mesh $\VV_p$ (we center the markers before regressing $\theta, \beta$ and $\rr$) and a free-form mesh $\VV_d$. Training losses ensure the consistency between $\VV$, $\VV_p$ and $\VV_d$.  We also show a detail of the cannonical marker placement, as attached on the GHUM model. Notice slight left/right and more pronounced front/back placement asymmetries that help disambiguate the model side and facing direction.
}
\label{fig:marker_poser}
\end{figure*}

Our approach is to use a 3d surface marker set as an intermediate representation proxy, controlled by both surface (mesh) properties and the parameters of a statistical 3d human pose and shape model (GHUM). For practical considerations, and without loss of generality, we adopt the Human3.6M marker set that consists of $N_m=53$ units, see fig.~\ref{fig:marker_poser} for details. We next describe two network heads, which given any 3d markers $\mathbf{M} \in \mathbb{R}^{N_m\times3}$ achieve the following: {\it (i)} reconstruct the GHUM mesh through a simple architecture $\VV_d(\mathbf{M}) \in \mathbb{R}^{N_v\times3}$, and {\it (ii)} recover the corresponding GHUM parameters $(\theta, \beta, \rr, \ttt)$ from $\mathbf{M}$, so we can also recover an anthropometric mesh equivalent to $\VV_d$, $\VV_p(\theta, \beta, \rr, \ttt)$.

\vspace{-4mm}

\paragraph{Training the Marker-based Poser (MP)}

The markers are essentially free 3 dof variables, but they follow the given surface placement description, in our case, the VICON protocol. To train a network that maps markers to vertices, we need a dataset of corresponding markers and vertices. 

We take a synthetic sampling approach based on our GHUM model. Given generative codes for pose and shape $\theta, \beta \in \mathcal{N}(\mathbf{0}; I)$, $\rr$ drawn from the Haar distribution on $SO(3)$, and $\ttt$ uniformly sampled from a $(-20 \ldots 20) \times (-20 \ldots 20) \times (-20 \ldots 20)$ meters box, we produce a posed GHUM  sample mesh $\mathbf{V}(\theta, \beta, \rr, \ttt)$. The associated markers can be retrieved by a simple (fixed) linear regression matrix $\WW \in \mathbb{R}^{N_v \times N_m} $, such that $\mathbf{M} = \WW\mathbf{V}(\theta, \beta, \rr, \ttt)$. In our experiments, we noticed that injecting noise at this point, \ie $\mathbf{M} + \mathcal{N}(\mathbf{0}; \epsilon I)$, supports the more accurate retrieval of the full mesh given real, imprecise markers that one could find in motion capture datasets such as CMU or Human3.6M, or as produced by an image-based marker regressor.
An overview of the poser function, denoted MP is given in fig. \ref{fig:marker_poser}. We denote $\mathbf{V}_d$ the mesh that is directly predicted from markers. We denote $\mathbf{V}_p$ the mesh that is parametrically obtained by posing the GHUM model given parameters $(\widetilde{\theta}, \widetilde{\beta}, \widetilde{\rr}, \widetilde{\ttt})$ regressed from markers. For training, we use the loss
\begin{align}
    \cL = \cL_{p}(\mathbf{V}, \mathbf{V}_p) + 
        \cL_{d}(\mathbf{V}, \mathbf{V}_d),
\end{align}
\noindent where $\cL_{P}$ and $\cL_{V}$ are the mean-per-vertex errors, computed using a $L_2$ metric, between the input mesh, and the parametric and direct meshes, respectively. We also experimented with supervising $(\widetilde{\theta}, \widetilde{\beta}, \widetilde{\rr}, \widetilde{\ttt})$ directly, but learning was not successful. 
%Our intuition is that in the normalizing flow pose prior there are many $\thetab$ configurations that correspond to the same marker positioning, so an inverse mapping is difficult to compute robustly. 
To make the training process easier, we subtract the mean marker position (computed as the 3d centroid of each $\MM$) before regressing $\theta, \beta$ and $\rr$ and we obtained lower reconstruction errors using this modification.

% \begin{figure}[!tbp]
% \begin{center}
% \includegraphics[width=.6\linewidth]{Figures/marker_placement.png}
% \end{center}
% \caption{\small Marker set. Front and back views as attached on the GHUM model. Notice slight left/right and more pronounced front/back placement asymmetries that help disambiguate the model side and facing direction.
% }
% \label{fig:marker_set}
% \end{figure}

\subsection {THUNDR}

\begin{figure*}[!ht]
\begin{center}
\includegraphics[width=0.9\linewidth]{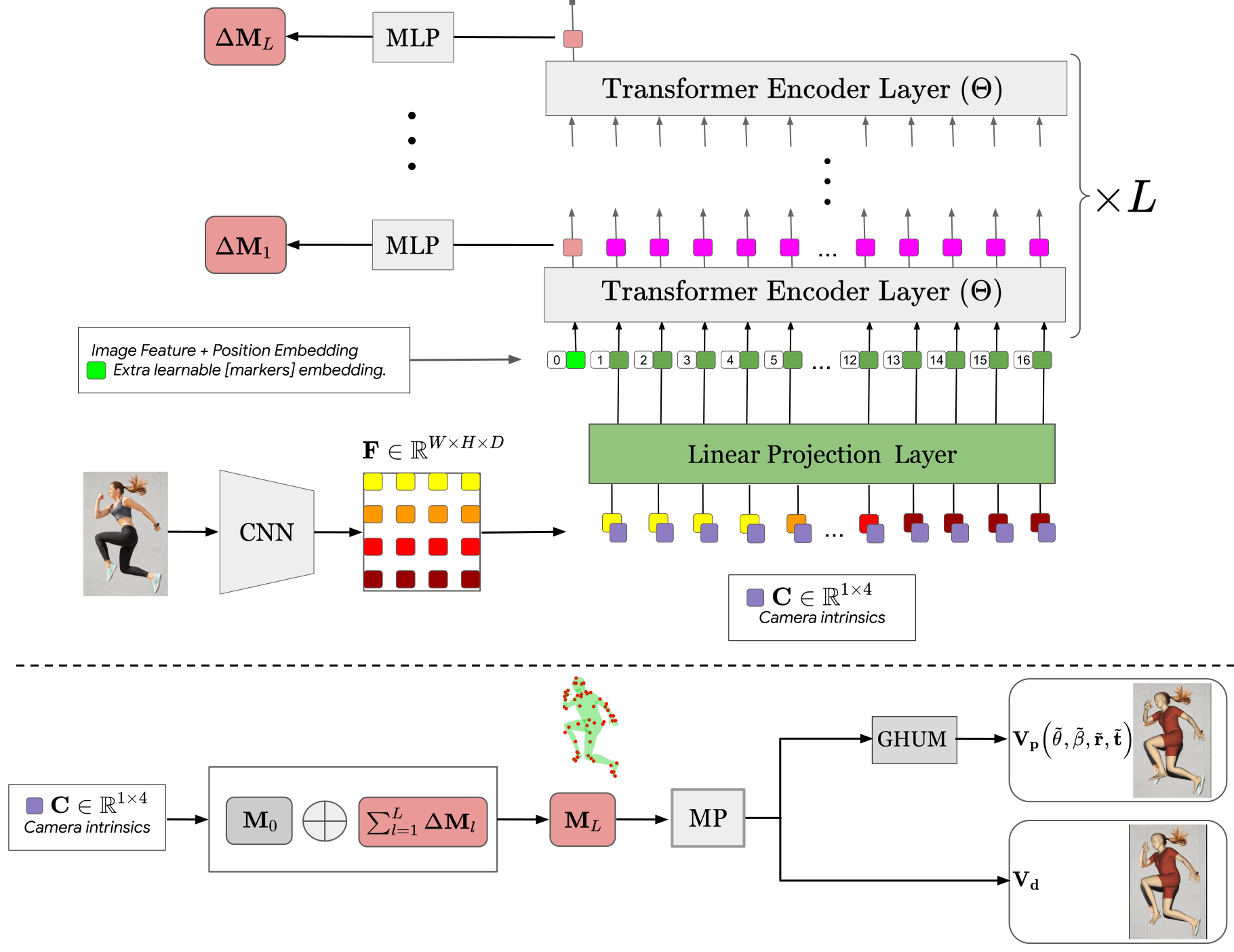}
\end{center}
\vspace{-5mm}
\caption{\small Overview of our proposed THUNDR architecture, to estimate the parameters of a generative human model (GHUM). \textit{(Top)} Given an input image, we first use a CNN to extract a feature map $\mathbf{F}\in \mathbb{R}^{W\times H\times D}$, where $W$ and $H$ represent the spatial extent, and $D$ the number of channels per feature. In this example $W=H=4$. We serialize the feature map and concatenate to each feature the camera intrinsics of the image, $\mathbf{C}$. Next, we take our sequence, linearly embed it and add positional encoding. We also add an extra learnable [\textit{markers}] token to the input. This representation is iteratively transformed $L$ times through \textit{the same} transformer encoder layer with learnable weights $\Theta$. At each transformation stage $l$, we gather the representation of the [\textit{markers}] token, feed it through an MLP and regress the marker coordinates refinement $\Delta \mathbf{M}_l$. \textit{(Bottom)} We compute the default marker coordinates $\mathbf{M}_0$ as a function of the image camera intrinsics and default GHUM model parameters. The regressed marker coordinates displacements are added to it and the result represents the final estimated marker coordinates $\mathbf{M}_L$. We use the pre-trained marker-based poser MP to get our predicted ghum model vertices and parameters.}
\label{fig:THUNDR}
\end{figure*}

In fig.~\ref{fig:THUNDR} we show an overview of our proposed hybrid learning architecture for monocular 3d body pose and shape estimation. Our architecture is different from existing pose and shape estimation methods, that directly regress the parameters of a human model (\ie SMPL or GHUM) from a single feature representation of an image. We instead regress an intermediate 3d representation in the form of surface landmarks (markers) and regularize it in training using a statistical body model. Moreover, we preserve the spatial structure of high-level image features by avoiding pooling operations, and relying instead on self-attention to enrich our representation~\cite{Vaswani2017}. We draw inspiration from vision transformers~\cite{dosovitskiy2020image}, as we also use a hybrid convolutional-transformer architecture, and from \cite{zanfir2020neural}, as we explore the idea of iteratively refining estimates by relying on cascaded, input-sensitive processing blocks, with homogeneous parameters. 

Our network receives as input a cropped image $\mathbf{I}\in\mathbb{R}^{W\times H\times3}$ of a person, together with the pseudo ground-truth camera intrinsics $\CC_c\in\mathbb{R}^{1\times4}$ of the crop (see \S~\ref{subsection:GHUM_CAMERA}). We apply a convolutional neural network (CNN) on the input image and extract a downsampled feature map representation $\mathbf{F}\in\mathbb{R}^{\frac{W}{32}\times \frac{H}{32}\times D}$. We flatten the feature map along the spatial dimensions to get a sequence of $N=\frac{W}{32}\times \frac{H}{32}$ tokens. We append to each token the camera intrinsics and get our input feature sequence $\mathbf{F}_s \in \mathbb{R}^{N\times(D+4)}$. This sequence is linearly embedded by means of matrix $\EE\in\mathbb{R}^{(D+4)\times D^\prime}$, where $D^\prime$ is the embedding dimensionality, and concatenate it with an extra learnable [\textit{markers}] token, $\mathbf{F}_{\text{m}} \in \mathbb{R}^{1\times D^\prime}$. Next, learnable positional embeddings $\mathbf{E}_{pos} \in \mathbb{R} ^ {(N+1)\times D^\prime}$ are added to the sequence. Different from standard transformer architectures, we use a single transformer encoder layer~\cite{Vaswani2017}, \textit{TL}, to iteratively refine our input representation for a number of $L$ steps. We collect at each stage $l\in{\{1\dots L}\}$, a refinement update $\Delta \mathbf{M}_l \in \mathbb{R}^{N_m\times3}$, with $N_m$ the number of markers, from each transformed representation $\mathbf{Z}_l$ using a shared MLP applied on the representation of the [\textit{markers}] token,  
%\begin{align}
%    \mathbf{Z}_0 &= [\mathbf{F}_{\text{m}}; \mathbf{F}_s\EE] + \EE_{pos} \nonumber\\
%    \mathbf{Z}_l &= \textit{TL}(\mathbf{Z}_{l-1})\nonumber\\
%    \Delta{\textbf{M}_l} &= \textit{MLP}(\mathbf{Z}_l^0)
%\end{align}
\begin{align}
    \mathbf{Z}_0 &=\begin{bmatrix}\mathbf{F}_{\text{m}} \\  \mathbf{F}_s\EE \end{bmatrix} + \EE_{pos} \\
    \mathbf{Z}_l &= \textit{TL}(\mathbf{Z}_{l-1})\\
    \Delta{\textbf{M}_l} &= \textit{MLP}(\mathbf{Z}_l^0).
\end{align}
The refinement updates $\Delta \mathbf{M}_l$ are added to the default marker coordinates, $\mathbf{M}_0$, as
\begin{equation}
    \mathbf{M}_L = \mathbf{M}_0 + \lambda \Sigma_{l=1}^L\Delta \mathbf{M}_l,
\end{equation}
\noindent where $\lambda$ is a parameter controlling the step size. $\mathbf{M}_0$ are computed based on the default GHUM parameters, $(\theta_0, \beta_0, \rr_0, \ttt_0)$, and camera intrinsics. That is, we find the optimal translation $\ttt_0^{*}$ such that the corresponding posed mesh projects in the center of the image~\cite{zanfir2020neural}. Finally, $\mathbf{M}_0$ is computed as

\begin{equation}
    \mathbf{M}_0 = \WW\mathbf{V}(\theta_0, \beta_0, \rr_0, \ttt_0^{*}).
\end{equation}

We apply the pre-trained marker-based poser MP (see \S~\ref{subsection:MARKER_POSER}) on $\mathbf{M}_L$ in order to recover the GHUM mesh and parameters, $\{\VV_d, \widetilde{\theta}, \widetilde{\beta}, \widetilde{\rr}, \widetilde{\ttt}\}$. We also compute the mesh geometry using the standard GHUM poser from the regressed model parameters, $\VV_p(\widetilde{\theta},\widetilde{\beta}, \widetilde{\rr}, \widetilde{\ttt})$. During training, we use a mixed regime based on both weak 2d supervision losses and full 3d supervision losses, where data is available.

We include regularization losses for pose and shape, as
\begin{equation}
     \mathcal{L}_{ps} = \|\widetilde{\beta}\|_2^2 + \|\widetilde{\theta}\|_2^2.
\end{equation}

For this constraint to also affect the predicted markers $\mathbf{M}_L$ in a direct manner, we must formulate a consistency loss between the two representations. We set a novel loss that measures the \textit{mean per-marker position error} (i.e. MPMPE) between the predicted markers and the markers on the surface of $\VV_p$, \ie $\MM_p = \WW \VV_p$, as
\begin{equation}
     \mathcal{L}_{m} = \frac{1}{N_m} \sum_{i=1}^{N_m}\|\mathbf{M}^i_L - \MM_p^i\|_2.
\end{equation}

We use a standard 2d reprojection loss measured with respect to either annotated or predicted keypoints, $\mathbf{j} \in \mathbb{R}^{K\times 2}$, weighted by a per-keypoint confidence score $\mathbf{s} \in \mathbb{R}^{K\times 1}$, with $K$ the number of keypoints. From our directly regressed mesh $\VV_d$ we extract 3d joints $\JJ$ via the standard GHUM regressor and project them using camera intrinsics $\CC_c$ to predict 2d keypoints
\begin{align}
    \mathcal{L}_{k} = \frac{1}{K} \sum_{i=1}^{K} \mathbf{s}_{i}\|\mathbf{j}_{i} - \Pi(\JJ_i(\VV_d), \CC_c)\|_2.
\end{align}

Similarly to \cite{zanfir2020neural}, we use a soft differentiable rasterizer ~\cite{liu2019soft} to compute a body part alignment loss with respect to either ground-truth or predicted body part maps $\BB \in \mathbb{R}^{W\times H\times 15}$, with $15$ different body part labels
\begin{align}
    \mathcal{L}_{b} = \frac{1}{W*H} \sum_{i=1}^{W*H} \|\BB_i - R(\VV_d, \CC_c)_i\|_1,
\end{align}

\noindent where $R$ is the rasterized image of the 3d body parts of $\VV_d$, projected using camera intrinsics $\CC_c$.

Given access to 3d supervision with ground-truth vertices $\VV_{gt}$ and joints $\JJ_{gt}$, we use standard vertex and 3d keypoints losses:
\begin{align}
\begin{split}
    \mathcal{L}_{f} = \lambda_{v} \mathcal{L}_{v}(\VV_d, \VV_{gt}) &+ \lambda_{j} \mathcal{L}_{j}(\JJ, \JJ_{gt})
    \notag,
\end{split}
\end{align}
with $\mathcal{L}_v$ the MPVE (mean per vertex error) metric and $\mathcal{L}_j$ the MPJPE (mean per joint position error) metric. Parameters $\lambda_{v}$ and $\lambda_{j}$ control the importance of each loss.

Finally, we can write our full loss function, as follows
\begin{equation}
    \mathcal{L} = \lambda_{ps}\mathcal{L}_{ps}+\lambda_m\mathcal{L}_m+\lambda_k\mathcal{L}_k +\lambda_b\mathcal{L}_b + \mathcal{L}_f
\end{equation}
where $\lambda$ are used to weigh the different loss components. The fully supervised loss $\mathcal{L}_f$ is only used if there exists 3d ground truth information.

\section{Experiments}\label{sec:exps}

\paragraph{Datasets}
We use two datasets containing images in-the-wild, COCO2017 \cite{lin2014microsoft} (30,000 images) and OpenImages \cite{OpenImages} (24,000 images) for our weakly-supervised training (WS). We use the 2d keypoint annotations where available, otherwise we rely on a 2d pose detector to supplement missing annotations and use an additional confidence score per keypoint.

For the fully-supervised (FS) experiments, we use two standard datasets Human3.6M~\cite{Ionescu14pami} and 3DPW~\cite{vonMarcard2018}. Because the ground-truth of 3DPW is provided as SMPL~\cite{SMPL2015} 3d meshes, we use GHUM fits to these meshes to report the vertex-to-vertex errors. The MPJPE metrics are reported on the 3d joints regressed from the ground-truth SMPL meshes, as standard in the literature. Differently from existing methods, we use less 3d supervision, with superior results. We did not include additional datasets such as MuCo-3DHP~\cite{mehta2018single}, MPI-INF-3DHP ~\cite{mehta2017vnect} or UP3D~\cite{Lassner2017UP3D}, but we believe they could be helpful in further increasing our reconstruction performance.

\vspace{-3mm}

\paragraph{Implementation details} In all our experiments we use a ResNet50~\cite{he2016identity} backbone pretrained for the ImageNet~\cite{imagenet} image classification task. 
%We did not experiment with using an HRNet~\cite{wang2020deep} backbone as in \cite{lin2020end}, where it shows that is has better performance than the ResNet50 variant. We leave this for future exploration.
Our complete architecture has $25$M parameters, $23.5$M for the backbone and $1.5$M for the transformer layer and the MLP regressor. We use $L=4$ stages, step size $\lambda = 0.1$, an embedding size $256$ and $8$ heads for the MultiHeadAttention layer. We train the network for $50$ epochs, with batch size of $32$, base learning rate of $1e-4$ and exponential decay $0.99$. Our marker poser MP has $8.5$M parameters and consists of MLPs with a hidden layer size of 256. The network is trained for $1$M steps with a batch size of $128$. All our networks were trained on a single V100 GPU with 16GB of memory. Our code is implemented in TensorFlow.

\begin{figure*}[!ht]
\begin{center}
\includegraphics[width=0.87\linewidth]{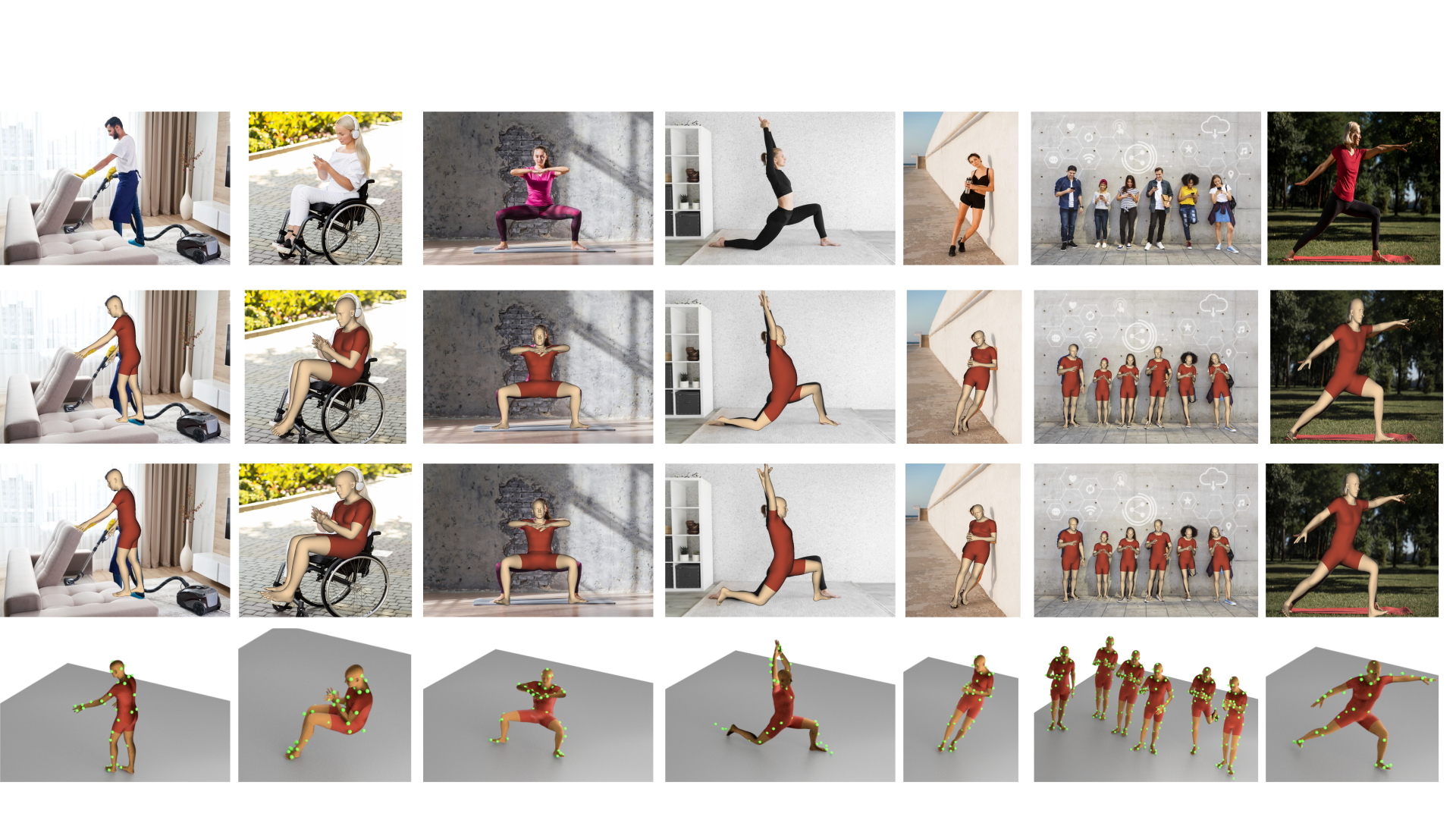}
\includegraphics[width=0.87\linewidth]{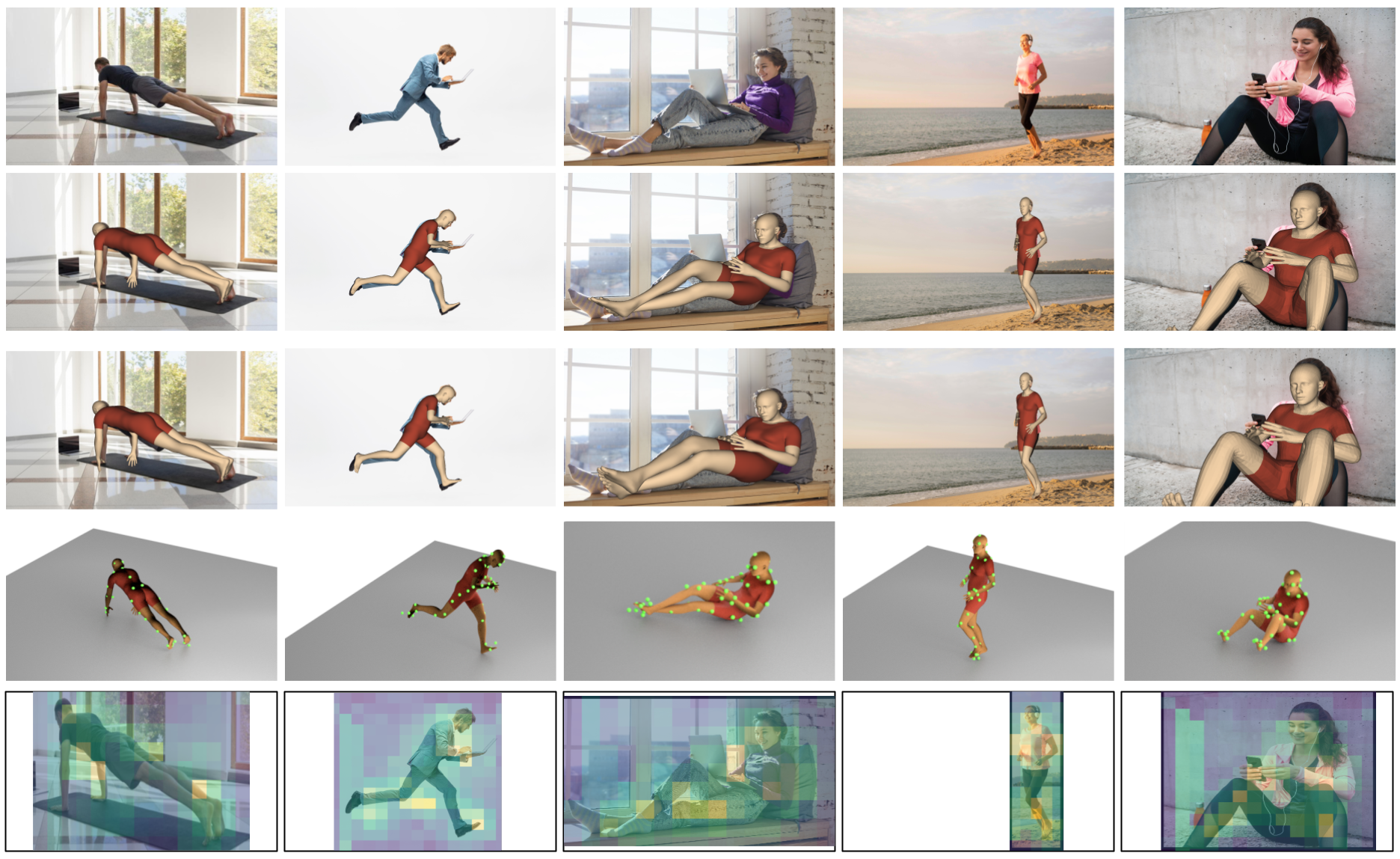}
\end{center}
\caption{\small Results of THUNDR on images in the wild. From top to bottom: {\it (i)} input image {\it (ii)} direct mesh reconstructions $\VV_d$ {\it (iii)} parameteric mesh reconstructions $\VV_p$. Notice that direct mesh reconstruction aligns better, particularly the feet and the limbs. {\it (iv)} reconstructions seen from a different viewpoint with regressed marker representations shown in green. {\it (v)} the image attention map for the [\textit{markers}] token, aggregated over all transformer layers. }
\label{fig:Results}
\vspace{-3mm}
\end{figure*}

\begin{table*}[!htbp]
    \small
    \centering
    \begin{tabular}[t]{|l||r|r|r|}
    \hline
    \textbf{Method}  & {MPJPE-PA} & {MPJPE} & {Translation Error} \\ 
    \hline
    \hline
    HMR (WS) \cite{Kanazawa2018} & $67.45$ & $106.84$ &NR \\
    \hline    
    HUND (SS) \cite{zanfir2020neural} & $66.0$ & $102$ & $175.0$\\
    \hline
    \textbf{THUNDR (WS)} & $\mathbf{62.2}$ & $\mathbf{87.0}$ & $161.6$ \\
    \hline
    \hline
    HMR \cite{Kanazawa2018} & $58.1$ & $88.0$& NR \\
    \hline
    HUND \cite{zanfir2020neural} & $53.0$ & $72.0$& $160.0$ \\
    \hline
    \textbf{THUNDR} & $\mathbf{39.8}$ & $\mathbf{55.0}$ & $\mathbf{143.9}$ \\
    \hline
    \end{tabular}
    \caption{\small Performance of different pose and shape estimation methods on the Human3.6M dataset, with training/testing under protocol P1.}
\label{tbl:H36MP1}
\end{table*}

\subsection{3D Pose and Shape Reconstruction}

For this task, we report several common error metrics that are used for
evaluating the error of 3d reconstruction. Most commonly used for 3d joint errors are mean per joint position error (MPJPE) and MJPE-PA, which is MPJPE after rigid alignment of the prediction with ground truth via Procrustes Analysis. The latter metric, removes global misalignment (\ie scale and rotation) and mainly evaluates the quality of the reconstructed 3d pose. For evaluating 3d shape we use the MPVPE metric between the vertices of the predicted and ground-truth meshes, respectively.

We evaluate our networks on the two datasets that provide 3d ground-truth information, Human3.6M and 3DPW. For the Human3.6M dataset, there are three commonly used evaluation protocols in the literature. Protocols P1 and P2 consider splitting the official training set into new training and testing subsets, with subjects S1, S5-S8 for training and S9 and S11 for evaluation. P1 evaluates on all available camera views in testing, while P2 only on a single predefined camera view (we consider this to be a highly inconclusive protocol due to its small size and design but report on it in order to compare to other methods). The third and most representative protocol we consider is the official one, where we evaluate on the hold-out test dataset of 900K samples. We also submit predictions on the official website for other methods (where code and models are available) to get comparable results. To be fair in our comparison with other methods, we do not retrain on the whole official training dataset.  We show results for all protocols in tables ~\ref{tbl:H36MP1}, \ref{tbl:H36MP2} and \ref{tbl:H36MOfficial}. For P1, we report results for both the weakly supervised regime (WS) and for the mixed regime (WS+FS) in order to compare with prior work. For the official protocol, we report only the MPJPE rounded to the nearest integer, as this is the format the results are returned by the official site. On all protocols and in all training regimes, we obtain state-of-the-art results. In table~\ref{tbl:3DPW} we report errors on the testing split of the 3DPW dataset. We obtain better results than the prior state-of-the-art, in both the WS+FS regime and the WS regime.

In fig.~\ref{fig:Results} we show qualitative reconstructions from THUNDR in-the-wild where one can observe that direct mesh reconstructions $\VV_d$ have better image alignment in general. We also show the image attention map for the [\textit{markers}] token, aggregated over all transformer layers. Notice how the network learns to focus on faces, hands and feet.
\vspace{-7mm}

\paragraph{Ablation Studies}
\begin{table}[!htbp]
    \small
    \centering
    \begin{tabular}[t]{|l||r|r|}
    \hline
    \textbf{Method}  & {MPJPE-PA} & {MPJPE}\\ 
    \hline
    THUNDR-GHUM (WS) & $63.5$ & $95.4$ \\
    \hline
    THUNDR-{$\VV_p$} (WS) & $61.8$ & $88.3$ \\
    \hline
    \textbf{THUNDR (WS)} & $\mathbf{59.7}$ & $\textbf{83.4}$ \\
    \hline
    \end{tabular}
    \vspace{-3mm}

    \caption{\small Ablation study on different variations of THUNDR: THUNDR-GHUM directly regresses GHUM parameters from the image and THUNDR-{$\VV_p$} is our standard version were we instead evaluate on the predicted parameteric mesh {$\VV_p$}. This evaluation is done in a weakly supervised regime and we report error metrics on Human3.6M protocol P2. }
\label{tbl:ablation_ghum}
\end{table}
In table~\ref{tbl:ablation_ghum}, we ablate different methodological choices in our proposed architecture in the weakly supervised regime and report results on protocol P2 of Human3.6M dataset. First, we change THUNDR to directly regress GHUM parameters (\ie $\widetilde{\beta}, \widetilde{\theta}, \widetilde{\rr}, \widetilde{\ttt}$) from the input image, skipping our intermediate marker representation and removing the marker poser MP. The convolutional-transformer architecture stays mostly the same, with some minor modifications to accommodate more output variables (\ie we use $4$ extra input tokens, one for each GHUM parameter, instead of $1$). This performs worse than our proposed architecture THUNDR and this shows that our intermediate marker representation is easier to learn from image features. Next, as in our full method we only use the direct mesh $\VV_d$, we also show the errors if we instead evaluate on the parameteric mesh $\VV_p$ (we denote this by THUNDR-$\VV_p$). These results are also better than THUNDR-GHUM, but worse than THUNDR. This again suggests the utility of our intermediate representation and the importance of working with two separate mesh reconstructions.

\vspace{-3mm}
\paragraph{Marker Poser}
%\paragraph{Marker Poser} 
We present more details on the training of the marker poser and its additional benefits, outside of the transformer-based 3d pose reconstruction architecture. During training, we experiment with $6$ levels of Gaussian noise added to the markers, as $\epsilon \in \{0, 2, 5, 10, 20, 50\}$ mm. We ablate each one of the trained marker poser models on the Human3.6M ground-truth marker data. The best performance in reconstruction is achieved for the network with $\epsilon = 5$ mm. This model is used in all of our other experiments. During training, the error on the direct mesh reconstruction reaches $2.5$ mm MPVPE, while the parameteric mesh reconstruction reaches $3.7$ mm MPVPE.
\vspace{-4mm}
\paragraph{Mesh Fitting} 
We test our marker-based poser on the Human3.6M dataset, for which the authors shared 3D marker positions for the training data. We fit an associated GHUM mesh in two ways: {\it (i)} by minimizing an energy that takes into account 3d marker ground truth, 2d reprojection errors for all GHUM 3d body joints (including hands and face) and a semantic alignment cost, and {\it ( ii)} by simply running our trained marker poser on the ground-truth 3d marker positions to produce a mesh $\VV_d$. For a sequence fitting example, see our Sup. Mat.
% \begin{figure*}[h]
% \begin{center}
% \includegraphics[width=1.\linewidth]{Figures/marker_fitting.png}
% \end{center}
% \caption{\small Deploying the marker poser on the GT markers of Human3.6M. \textit{Left} Image with markers. \textit{Middle-left} Non-linear optimization-based marker fitting. \textit{Middle-right} Our $\VV_d$ output. \textit{Right} Our $\VV_p$ output.
% }
% \label{fig:marker_fitting}
% \end{figure*}
 First, we compute the mean per-marker error for the models $\VV_{gt}$ obtained from energy optimization to ground-truth makers $\MM_{gt}$ (\ie those recovered from mocap data). This gives an error of $38.4$ mm, with an average processing rate of $0.15$ frames/second. Second, we compute the errors of markers placed on the predicted mesh $\VV_d$ given ground-truth marker positions. This achieves a slightly higher error of $44.3$ mm, but with an average processing rate of $100$ frames/second, when ran sequentially. Note that our marker poser has never seen the marker sequences of Human3.6M during training, as the marker poser was trained with samples drawn from a normalizing flow prior based on the CMU motion capture dataset~\cite{CMUMocap}.%Note that these errors assume that the mocap markers were placed and tracked correctly during their respective recording sessions, for a meaningful comparison.
\begin{table}[!htbp]
    \small
    \centering
    \begin{tabular}[t]{|l||r|r|}
    \hline
    \textbf{Method}  & {MPJPE-PA} & {MPJPE} \\ 
    \hline
    \hline
    HMR \cite{Kanazawa2018} & $56.8$ & $88.0$\\
    \hline
    GraphCMR \cite{kolotouros2019convolutional} & $50.1$ &NR \\
    \hline
    Pose2Mesh \cite{choi2020pose2mesh} & $47.0$ &$64.9$ \\
    \hline
    I2L-MeshNet \cite{moon2020MeshNet} & $41.7$ & $55.7$ \\
    %\hline
    %Vibe \cite{kocabas2019vibe} & $41.4$ & $65.6$ \\
    \hline
    SPIN \cite{kolotouros2019learning} & $41.1$ &NR\\
    %\hline
    %METRO (standard backbone) \cite{lin2020end} & $40.6$ & $56.5$\\
    %\hline
    %METRO \cite{lin2020end} & $36.7$ & $54.0$\\
    \hline
    \hline
    \textbf{THUNDR} & $\mathbf{34.9}$ & $\mathbf{48.0}$  \\
    \hline
    \end{tabular}
    \vspace{-3mm}
    \caption{\small Performance of different pose and shape estimation methods on the Human3.6M dataset, protocol P2.}
\label{tbl:H36MP2}
\end{table}

\begin{table}[!htbp]
    \small
    \centering
    \begin{tabular}[t]{|l||r|r|r|}
    \hline
    \textbf{Method}  & {MPJPE-PA} & {MPJPE} & {MPVPE} \\ 
    \hline
    \hline
    HUND \cite{zanfir2020neural} (SS) & $70.3$ & $98.1$ & NR\\
    \textbf{THUNDR (WS)} & $\mathbf{59.9}$ & $\mathbf{86.8}$ &NR \\
    \hline
    \hline
    HMR \cite{Kanazawa2018} & $81.3$ & NR &NR\\
    \hline
    GraphCMR \cite{kolotouros2019convolutional} & $70.2$ &NR &NR\\
    \hline
    SPIN \cite{kolotouros2019learning} & $59.2$ &NR & $116.4$\\
    \hline
    Pose2Mesh \cite{choi2020pose2mesh} & $58.9$ &$89.2$ &NR\\
    \hline
    I2L-MeshNet \cite{moon2020MeshNet} & $57.7$ & $93.2$ &NR\\
    \hline
    HUND \cite{zanfir2020neural} & $56.5$ & $87.7$ & NR\\
    %\hline
    %Vibe \cite{kocabas2019vibe} & $51.9$ & $82.0$ & $99.1$\\
    %\hline
    %METRO \cite{lin2020end} & $\mathbf{47.9}$ & $77.1$ & $88.2$\\
    \hline
    %\textbf{THUNDR} & $54.0$ & $\mathbf{76.0}$ &NR \\
    \textbf{THUNDR} & $\mathbf{51.5}$ & $\mathbf{74.8}$ & *$\mathbf{88.0}$ \\
    \hline
    \end{tabular}
    \vspace{-3mm}
    \caption{\small Performance of different pose and shape estimation methods on the 3DPW dataset.*\textit{Shape evaluation is done on GHUM.}}
\label{tbl:3DPW}
\end{table}

\begin{table}[!htbp]
    \small
    \centering
    \begin{tabular}[t]{|l||r|}
    \hline
    \textbf{Method}  & {MPJPE}\\ 
    \hline
    \hline
    HMR \cite{Kanazawa2018} & $89$\\
    \hline
    SPIN \cite{kolotouros2019learning} & $68$ \\
    \hline
    HUND \cite{zanfir2020neural} & ${66}$ \\
    \hline
    \textbf{THUNDR} & $\mathbf{53}$ \\
    \hline
    \end{tabular}
    \vspace{-3mm}
    \caption{\small Performance of different methods on the Human3.6M official, representative held-out test set, containing 900K samples. }
\label{tbl:H36MOfficial}
\end{table}

\vspace{-4mm}
\paragraph{Ethical Considerations} Our methodology aims to decrease bias by introducing flexible forms of self-supervision which would allow, in principle, for system bootstrapping and adaptation to new domains and fair, diverse subject distributions, for which labeled data may be difficult or impossible to collect upfront. Applications like visual surveillance and person identification would not be effectively supported currently, given that model's output does not provide sufficient detail for these purposes. This is equally true of the creation of potentially adversely-impacting deepfakes, as we do not include an appearance model or a joint audio-visual model.

\section{Conclusions}

We have presented THUNDR, a transformer-based deep neural network methodology to reconstruct the 3d pose and shape of people, given monocular RGB images.  Faced with the difficult issue of handling not directly observable human body joints, on which nevertheless many error metrics are based, and aiming at both reconstruction accuracy and good self-supervised learning and generalization under anthropometric human body constraints, we propose a novel model that combines a surface-marker representation with 3d statistical body regularization. The model is designed around a learnable pipeline that refines multiple transformer layers for computational efficiency and for precise, task-sensitive, image feature localization.
%As a by-product, we show how marker data can be easily converted into a complete parametric human model (in our case GHUM), by employing a novel deep learning architecture. 
We demonstrate state-of-the-art results on Human3.6M and 3DPW, in both the fully-supervised and the self-supervised regimes. In our extensive qualitative assessment (see Sup. Mat.) we observe  accurate 3d reconstruction performance for difficult human poses collected under challenging imaging conditions.

%As a by-product, we show how marker data can be easily converted into a complete parametric human model (in our case GHUM), by employing a novel deep learning architecture. 
%We demonstrate state-of-the-art results on Human3.6M and 3DPW, in both the fully-supervised and the self-supervised regimes. In our extensive qualitative assessment (see Sup. Mat.) we observe  accurate 3d reconstruction performance for difficult human poses collected under challenging imaging conditions.

\clearpage
\bibliographystyle{ieee_fullname}
\bibliography{egbib}

\begin{thebibliography}{10}\itemsep=-1pt

\bibitem{CMUMocap}
{Carnegie Mellon Motion Capture Database. http://mocap.cs.cmu.edu. }.

\bibitem{arnab2019exploiting}
Anurag Arnab, Carl Doersch, and Andrew Zisserman.
\newblock Exploiting temporal context for 3d human pose estimation in the wild.
\newblock In {\em CVPR}, pages 3395--3404, 2019.

\bibitem{biggs20203d}
Benjamin Biggs, S{\'e}bastien Ehrhadt, Hanbyul Joo, Benjamin Graham, Andrea
  Vedaldi, and David Novotny.
\newblock 3d multi-bodies: Fitting sets of plausible 3d human models to
  ambiguous image data.
\newblock {\em arXiv preprint arXiv:2011.00980}, 2020.

\bibitem{bogo2016}
Federica Bogo, Angjoo Kanazawa, Christoph Lassner, Peter Gehler, Javier Romero,
  and Michael~J Black.
\newblock Keep it {SMPL}: Automatic estimation of 3d human pose and shape from
  a single image.
\newblock In {\em ECCV}, 2016.

\bibitem{choi2020pose2mesh}
Hongsuk Choi, Gyeongsik Moon, and Kyoung~Mu Lee.
\newblock Pose2mesh: Graph convolutional network for 3d human pose and mesh
  recovery from a 2d human pose.
\newblock In {\em ECCV}, pages 769--787. Springer, 2020.

\bibitem{ExPose:2020}
Vasileios Choutas, Georgios Pavlakos, Timo Bolkart, Dimitrios Tzionas, and
  Michael~J. Black.
\newblock Monocular expressive body regression through body-driven attention.
\newblock In {\em European Conference on Computer Vision (ECCV)}, 2020.

\bibitem{imagenet}
J. Deng, W. Dong, R. Socher, L.-J. Li, K. Li, and L. Fei-Fei.
\newblock {ImageNet: A Large-Scale Hierarchical Image Database}.
\newblock In {\em CVPR}, 2009.

\bibitem{dosovitskiy2020image}
Alexey Dosovitskiy, Lucas Beyer, Alexander Kolesnikov, Dirk Weissenborn,
  Xiaohua Zhai, Thomas Unterthiner, Mostafa Dehghani, Matthias Minderer, Georg
  Heigold, Sylvain Gelly, et~al.
\newblock An image is worth 16x16 words: Transformers for image recognition at
  scale.
\newblock {\em arXiv preprint arXiv:2010.11929}, 2020.

\bibitem{georgakis2020hierarchical}
Georgios Georgakis, Ren Li, Srikrishna Karanam, Terrence Chen, Jana
  Ko{\v{s}}eck{\'a}, and Ziyan Wu.
\newblock Hierarchical kinematic human mesh recovery.
\newblock In {\em ECCV}, pages 768--784. Springer, 2020.

\bibitem{guler2019holopose}
Riza~Alp Guler and Iasonas Kokkinos.
\newblock Holopose: Holistic 3d human reconstruction in-the-wild.
\newblock In {\em CVPR}, pages 10884--10894, 2019.

\bibitem{he2016identity}
Kaiming He, Xiangyu Zhang, Shaoqing Ren, and Jian Sun.
\newblock Identity mappings in deep residual networks.
\newblock In {\em ECCV}, pages 630--645. Springer, 2016.

\bibitem{Ionescu14pami}
C. Ionescu, D. Papava, V. Olaru, and C. Sminchisescu.
\newblock Human3.6{M}: Large scale datasets and predictive methods for 3d human
  sensing in natural environments.
\newblock {\em PAMI}, 2014.

\bibitem{iqbal2020weakly}
Umar Iqbal, Pavlo Molchanov, and Jan Kautz.
\newblock Weakly-supervised 3d human pose learning via multi-view images in the
  wild.
\newblock In {\em CVPR}, pages 5243--5252, 2020.

\bibitem{jiang2020coherent}
Wen Jiang, Nikos Kolotouros, Georgios Pavlakos, Xiaowei Zhou, and Kostas
  Daniilidis.
\newblock Coherent reconstruction of multiple humans from a single image.
\newblock In {\em CVPR}, pages 5579--5588, 2020.

\bibitem{Kanazawa2018}
Angjoo Kanazawa, Michael~J. Black, David~W. Jacobs, and Jitendra Malik.
\newblock End-to-end recovery of human shape and pose.
\newblock In {\em CVPR}, 2018.

\bibitem{kolotouros2019learning}
Nikos Kolotouros, Georgios Pavlakos, Michael~J Black, and Kostas Daniilidis.
\newblock Learning to reconstruct 3d human pose and shape via model-fitting in
  the loop.
\newblock In {\em Proceedings of the IEEE International Conference on Computer
  Vision}, pages 2252--2261, 2019.

\bibitem{kolotouros2019convolutional}
Nikos Kolotouros, Georgios Pavlakos, and Kostas Daniilidis.
\newblock Convolutional mesh regression for single-image human shape
  reconstruction.
\newblock In {\em Proceedings of the IEEE Conference on Computer Vision and
  Pattern Recognition}, pages 4501--4510, 2019.

\bibitem{OpenImages}
Alina Kuznetsova, Hassan Rom, Neil Alldrin, Jasper Uijlings, Ivan Krasin, Jordi
  Pont-Tuset, Shahab Kamali, Stefan Popov, Matteo Malloci, Tom Duerig, and
  Vittorio Ferrari.
\newblock The open images dataset v4: Unified image classification, object
  detection, and visual relationship detection at scale.
\newblock {\em arXiv:1811.00982}, 2018.

\bibitem{Lassner2017UP3D}
Christoph Lassner, Javier Romero, Martin Kiefel, Federica Bogo, Michael~J.
  Black, and Peter~V. Gehler.
\newblock Unite the people: Closing the loop between 3d and 2d human
  representations.
\newblock In {\em IEEE Conf. on Computer Vision and Pattern Recognition
  (CVPR)}, July 2017.

\bibitem{lin2014microsoft}
Tsung-Yi Lin, Michael Maire, Serge Belongie, James Hays, Pietro Perona, Deva
  Ramanan, Piotr Doll{\'a}r, and C~Lawrence Zitnick.
\newblock Microsoft coco: Common objects in context.
\newblock In {\em ECCV}, 2014.

\bibitem{liu2019soft}
Shichen Liu, Tianye Li, Weikai Chen, and Hao Li.
\newblock Soft rasterizer: A differentiable renderer for image-based 3d
  reasoning.
\newblock In {\em ICCV}, pages 7708--7717, 2019.

\bibitem{SMPL2015}
Matthew Loper, Naureen Mahmood, Javier Romero, Gerard Pons-Moll, and Michael~J.
  Black.
\newblock {SMPL}: A skinned multi-person linear model.
\newblock {\em SIGGRAPH}, 2015.

\bibitem{mehta2018single}
Dushyant Mehta, Oleksandr Sotnychenko, Franziska Mueller, Weipeng Xu, Srinath
  Sridhar, Gerard Pons-Moll, and Christian Theobalt.
\newblock Single-shot multi-person 3d pose estimation from monocular rgb.
\newblock In {\em 3DV}, pages 120--130. IEEE, 2018.

\bibitem{mehta2017vnect}
Dushyant Mehta, Srinath Sridhar, Oleksandr Sotnychenko, Helge Rhodin, Mohammad
  Shafiei, Hans-Peter Seidel, Weipeng Xu, Dan Casas, and Christian Theobalt.
\newblock Vnect: Real-time 3d human pose estimation with a single rgb camera.
\newblock {\em ACM Transactions on Graphics (TOG)}, 2017.

\bibitem{moon2020MeshNet}
Gyeongsik Moon and Kyoung~Mu Lee.
\newblock I2l-meshnet: Image-to-lixel prediction network for accurate 3d human
  pose and mesh estimation from a single rgb image.
\newblock In {\em European Conference on Computer Vision (ECCV)}, 2020.

\bibitem{dmhs_cvpr17}
A.I. Popa, M. Zanfir, and C. Sminchisescu.
\newblock {Deep Multitask Architecture for Integrated 2D and 3D Human Sensing}.
\newblock In {\em CVPR}, 2017.

\bibitem{Rhodin_2018_ECCV}
Helge Rhodin, Mathieu Salzmann, and Pascal Fua.
\newblock Unsupervised geometry-aware representation for 3d human pose
  estimation.
\newblock In {\em ECCV}, September 2018.

\bibitem{sminchisescu_ijrr03}
C. Sminchisescu and B. Triggs.
\newblock {Estimating Articulated Human Motion with Covariance Scaled
  Sampling}.
\newblock {\em IJRR}, 22(6):371--393, 2003.

\bibitem{sun2018integral}
Xiao Sun, Bin Xiao, Fangyin Wei, Shuang Liang, and Yichen Wei.
\newblock Integral human pose regression.
\newblock In {\em Proceedings of the European Conference on Computer Vision
  (ECCV)}, pages 529--545, 2018.

\bibitem{varol18_bodynet}
G{\"u}l Varol, Duygu Ceylan, Bryan Russell, Jimei Yang, Ersin Yumer, Ivan
  Laptev, and Cordelia Schmid.
\newblock {BodyNet}: Volumetric inference of {3D} human body shapes.
\newblock In {\em ECCV}, 2018.

\bibitem{Vaswani2017}
Ashish Vaswani, Noam Shazeer, Niki Parmar, Jakob Uszkoreit, Llion Jones,
  Aidan~N. Gomez, Lukasz Kaiser, and Illia Polosukhin.
\newblock Attention is all you need.
\newblock In {\em NIPS}, 2017.

\bibitem{vonMarcard2018}
Timo von Marcard, Roberto Henschel, Michael Black, Bodo Rosenhahn, and Gerard
  Pons-Moll.
\newblock Recovering accurate 3d human pose in the wild using {IMUs} and a
  moving camera.
\newblock In {\em ECCV}, 2018.

\bibitem{ghum2020}
Hongyi Xu, Eduard~Gabriel Bazavan, Andrei Zanfir, Bill Freeman, Rahul
  Sukthankar, and Cristian Sminchisescu.
\newblock {GHUM} \& {GHUML}: Generative {3D} human shape and articulated pose
  models.
\newblock {\em CVPR}, 2020.

\bibitem{xu2019denserac}
Yuanlu Xu, Song-Chun Zhu, and Tony Tung.
\newblock Denserac: Joint 3d pose and shape estimation by dense
  render-and-compare.
\newblock In {\em Proceedings of the IEEE International Conference on Computer
  Vision}, pages 7760--7770, 2019.

\bibitem{zanfir2020weakly}
Andrei Zanfir, Eduard~Gabriel Bazavan, Hongyi Xu, Bill Freeman, Rahul
  Sukthankar, and Cristian Sminchisescu.
\newblock Weakly supervised 3d human pose and shape reconstruction with
  normalizing flows.
\newblock {\em ECCV}, 2020.

\bibitem{zanfir2020neural}
Andrei Zanfir, Eduard~Gabriel Bazavan, Mihai Zanfir, William~T Freeman, Rahul
  Sukthankar, and Cristian Sminchisescu.
\newblock Neural descent for visual 3d human pose and shape.
\newblock {\em CVPR 2021, arXiv preprint arXiv:2008.06910}.

\bibitem{zanfir2018monocular}
Andrei Zanfir, Elisabeta Marinoiu, and Cristian Sminchisescu.
\newblock Monocular 3d pose and shape estimation of multiple people in natural
  scenes-the importance of multiple scene constraints.
\newblock In {\em CVPR}, 2018.

\bibitem{zeng20203d}
Wang Zeng, Wanli Ouyang, Ping Luo, Wentao Liu, and Xiaogang Wang.
\newblock 3d human mesh regression with dense correspondence.
\newblock In {\em CVPR}, pages 7054--7063, 2020.

\bibitem{zhang2020object}
Tianshu Zhang, Buzhen Huang, and Yangang Wang.
\newblock Object-occluded human shape and pose estimation from a single color
  image.
\newblock In {\em CVPR}, pages 7376--7385, 2020.

\bibitem{zhou2018continuity}
Yi Zhou, Connelly Barnes, Jingwan Lu, Jimei Yang, and Hao Li.
\newblock On the continuity of rotation representations in neural networks.
\newblock {\em arXiv preprint arXiv:1812.07035}, 2018.

\end{thebibliography}

\end{document}